\documentclass[conference]{IEEEtran}
\IEEEoverridecommandlockouts
\usepackage{cite}
\usepackage{amsmath,amssymb,amsfonts}
\usepackage{algorithmic}
\usepackage{graphicx}
\usepackage{textcomp}
\usepackage{xcolor}
\usepackage{booktabs}
\usepackage{acronym}
\newacro{CNN}[CNN]{Convolutional Neural Network}
\newacro{BoF}[BoF]{Bag of Features}

\def\BibTeX{{\rm B\kern-.05em{\sc i\kern-.025em b}\kern-.08em
    T\kern-.1667em\lower.7ex\hbox{E}\kern-.125emX}}
\begin{document}

\title{Efficient CNN with uncorrelated Bag of Features pooling
\thanks{This  work  was  supported  by   NSF-Business Finland  Center  for Visual and Decision Informatics (CVDI) project AMALIA. }
}
\author{\IEEEauthorblockN{Firas Laakom}
\IEEEauthorblockA{\textit{Faculty of Information Technology and Communication Sciences} \\
\textit{ Tampere University}\\
Tampere, Finland \\
 firas.laakom@tuni.fi}
\and 
\IEEEauthorblockN{Jenni Raitoharju}
\IEEEauthorblockA{\textit{Faculty of Information Technology} \\
\textit{University of Jyväskylä}\\
Jyväskylä, Finland\\
jenni.k.raitoharju@jyu.fi}
\and
\IEEEauthorblockN{ Alexandros Iosifidis}
\IEEEauthorblockA{\textit{Department of Electrical and Computer Engineering} \\
\textit{Aarhus University }\\
Aarhus, Denmark \\
ai@ece.au.dk}

\and

\IEEEauthorblockN{ Moncef Gabbouj}
\IEEEauthorblockA{\textit{Faculty of Information Technology and Communication Sciences} \\
\textit{ Tampere University}\\
Tampere, Finland \\
moncef.gabbouj@tuni.fi}
}

\maketitle

\begin{abstract}
Despite the superior performance of CNN,  deploying them on low computational power devices is still limited as they are typically computationally expensive. One key cause of the high complexity is the connection between the convolution layers and the fully connected layers, which typically requires a high number of parameters. To alleviate this issue, Bag of Features (BoF) pooling has been recently proposed. BoF learns a dictionary, that is used to compile a histogram representation of the input. In this paper, we propose an approach that builds on top of BoF pooling to boost its efficiency by ensuring that the items of the learned dictionary are non-redundant. We propose an additional loss term, based on the pair-wise correlation of the items of the dictionary, which complements the standard loss to explicitly regularize the model to learn a more diverse and rich dictionary. The proposed strategy yields an efficient variant of BoF and further boosts its performance, without any additional parameters.
\end{abstract}


\section{Introduction}
In recent years, \acp{CNN} have significantly advanced many tasks in the computer vision field due to their ability to learn `good' feature representation in an end-to-end manner \cite{lecun2015deep,girshick2015fast}. However, despite their superior performance across multiple tasks, e.g., image classification \cite{krizhevsky2012imagenet,he2016deep,simonyan2014very}, object detection \cite{ren2015faster, zhao2019object,tan2020efficientdet}, anomaly detection \cite{pang2021deep,ruff2018deep,park2020learning,kim2018encoding},  deploying \ac{CNN}-based solutions on low computational power devices, such as mobile phones, is still limited as most of the high-accuracy models are typically computationally expensive \cite{lecun2015deep,8763885,laakom2019color}. Thus, they are inefficient in terms of time and energy consumption \cite{goodfellow2016deep}. To alleviate this issue, several approaches have been proposed to reduce the number of parameters required by a CNN model \cite{Li_2019_CVPR,lin2016towards,bof,bofp,menghani2021efficient}.

The standard CNN model is usually composed of two parts: The first part is formed of convolutional layers typically coupled with max-pooling operations. Then, in the second part,  fully connected layers are connected directly to a flattened version of the last convolutional layer output. This connection dramatically increases the total number of parameters, as convolutional layer outputs usually have high dimensionality. Recent approaches mitigate this problem by developing better mechanisms for connecting both parts, e.g., global average pooling and \ac{BoF}  pooling.

\begin{figure}[t]
\centering
\includegraphics[width=0.97\linewidth]{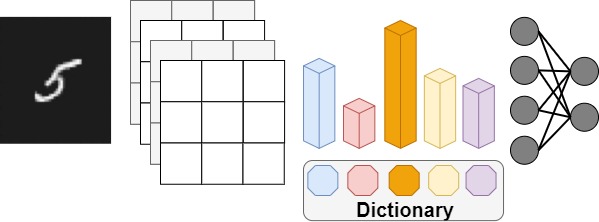}
\caption{An illustration of a simple \ac{BoF}-based CNN model. From left to right: Input image, convolutional layer, \ac{BoF} layer containing a dictionary and outputting a histogram, and fully connected layers.}
\label{illustration}
\end{figure}

\ac{BoF} pooling is a neural extension \cite{bofp,bof} of the famous  Bag-of-Visual Words \cite{boforigi, boforigi2,yang2007evaluating,csurka2010fisher,shukla2022bag,zhang2010understanding}. An illustration of a simple \ac{BoF}-based CNN model is presented in Figure \ref{illustration}. Based on the convolutional output, \ac{BoF} pooling learns a codebook (dictionary) and outputs a shallow histogram representation of the input. The items of the dictionary are optimized in an end-to-end manner during the standard back-propagation. This yields powerful and efficient models, that achieve a high performance with a low computational footprint. Recently, \ac{CNN} models, based on \ac{BoF} pooling, have been used to solve multiple tasks \cite{passalis2018training,chumachenko2022self,8487014,bof3, passalis2020efficient,tran2020attention}, such as action recognition \cite{bof3}, information retrieval \cite{bof2}, and illumination estimation \cite{laakom2020bag}.

In this paper, we propose an approach that builds on top of \ac{BoF} pooling to boost its efficiency by ensuring that the items of the learned dictionary are non-redundant. Forcing an uncorrelated structure on the codebook yields a more powerful model which can achieve a high performance with minimal dictionary size. Diversity in deep learning context has been shown to lead to better results \cite{chen2008diversity,gangeh2015supervised,rajesh2018heuristic,ayinde2019regularizing,naderahmadian2015correlation,van2012kernel,jing2014uncorrelated,wang2020orthogonal,zbontar2021barlow,pmlrxie17b,cogswell2015reducing,laakom2021feature,laakom2022reducing,laakom2021within}. To this end, we propose to augment the loss of the model to penalize pair-wise correlations between the items of the codebook. The proposed technique requires no additional parameters and can be incorporated in any \ac{BoF}-based CNN model to boost the performance of the CNN model. 

The contributions of this paper can be summarized as follows: 
\begin{itemize}
    \item We propose a scheme to avoid redundant items in the dictionary learned by the \ac{BoF}.
    \item We propose to augment the CNN-loss to explicitly penalize the pair-wise correlations between codebook items and learn rich compressed dictionary.
    \item The proposed regularizer acts as an unsupervised regularizer on top of the \ac{BoF} pooling layer and can be integrated into any \ac{BoF}-based CNN model in a plug-and-play manner.
    \item The proposed approach is evaluated with three datasets. The results show a consistent performance boost compared to the standard approach.
\end{itemize}

The rest of this paper is organized as follows. First, provide a brief overview of \ac{BoF} in Section \ref{bofsec}. In Section \ref{propappr}, we present the proposed  approach. In Section \ref{experimentalresults}, we empirically evaluate the performance of our method on three different datasets. We conclude the paper in Section \ref{concfut}.

\section{Bag-of-Features Pooling} \label{bofsec}

In this section, we briefly describe the \ac{BoF} pooling mechanism. \ac{BoF} \cite{bof,bofp} has been incorporated in a variety of applications and often led to superior results \cite{bof2,laakom2020bag,passalis2020efficient}. The \ac{BoF} pooling is parameterized with a dictionary. Given an input, i.e., the output maps of the last convolutional layer,  a histogram representation is compiled based on the dictionary. In the training phase, the items of the dictionary are optimized with the traditional back-propagation. The size of the dictionary is a hyper-parameter that can be adjusted with a validation set to avoid over-fitting. 

BoF pooling is formed using two inner layers: a Radial Basis Function (RBF) layer that measures the similarity of the input features to the RBF centers and an accumulation layer that builds a histogram of the quantized feature vectors. Formally, let \textbf{X} be the input image and $\rho(\textbf{X}) \in \mathbb{R}^{D \times P}$ the output of the convolutional layer, the RBF layer outputs a sequence of of quantized representations: 
\begin{equation*}
\Psi = [ \psi_1,  \psi_2, \cdots, \psi_P]\in \mathbb{R}^{K \times P},
\end{equation*}
where $\psi_i$ is the representation corresponding to the $i^{th}$ feature, i.e.,
\begin{equation*}
\psi_i = [\psi_{i,1}, \cdots, \psi_{i,K}].
\end{equation*}

The output of the $i^{th}$ RBF unit is as follows:
\begin{equation}
            \psi_{n,i} = \frac{\exp(- || \rho(\textbf{X})_n - \textbf{c}_i || / m_i )}{ \sum_j \exp(- || \rho(\textbf{X})_n - \textbf{c}_j || / m_j )  }  ,
\end{equation}
where $\textbf{c}_i$ is the center of the i-th RBF neuron,  and $m_i$ is a scaling factor. The outputs of the $P$ RBF neurons are accumulated in the next layer in order to obtain the final representation $\Phi$ of each image: 
\begin{equation}
          \Phi = \frac{1}{P} \sum_j \psi_j.
\end{equation}

To summarize, \ac{BoF} receives as input a feature representation, usually in high dimension, and quantizes it into a fixed-size shallow histogram representation. The quantization is based on the inner dictionary, $\{\textbf{c}_1, \cdots, \textbf{c}_K\}$, which can be learned jointly with the rest of the parameters in an end-to-end manner.

\section{Our Approach} \label{propappr}
\ac{BoF} pooling layer is a key technique that can be used in \acp{CNN} to construct powerful models with a low computational cost. The \ac{BoF} relies on a dictionary, learned during the training, to compute its shallow output. In this paper, we propose an approach that builds on top of \ac{BoF} pooling to boost its efficiency by explicitly forcing the items of the learned dictionary to be distinct and non-redundant. We propose a simple additional regularizer that penalizes the similarities between the codebook items.  This can further boost the performance of the model, without any additional parameters. 

\begin{figure*}[t]
\centering
\includegraphics[width=0.7\linewidth]{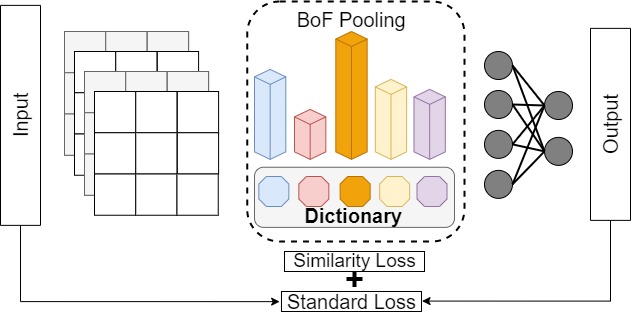}
\caption{An illustration on how the \ac{BoF}-based CNN model loss is computed using our approach. The standard loss can be least squares or cross entropy and the similarity loss corresponds to the second term in \eqref{loss_eq}. }
\label{illustration_loss}
\end{figure*}

The dictionary learned the \ac{BoF} layer plays a critical role in the global performance of the model. Intuitively, Learning a diverse and rich dictionary yields in a robust codebook and increases the efficiency of the global model. Given a CNN model containing a \ac{BoF} pooling layer with an inner dictionary $\{ \textbf{c}_1,\cdots,\textbf{c}_K \}$ of size $K$, the similarity $SIM$
 between two elements $\textbf{c}_i$ and $\textbf{c}_j$ of this dictionary can be measured with the squared  correlation:
\begin{equation} 
SIM(\textbf{c}_i,\textbf{c}_j ) = \Big( corr(\textbf{c}_i,\textbf{c}_j ) ) \Big)^2,
\end{equation}
where $ corr(\cdot,\cdot) $ is the correlation operator.

Intuitively, $SIM$ measures how similar two items are. Our goal is to regularize the similarities between the elements of the dictionary. So, the global similarity regularizer can be computed as the the sum of the pair-wise similarities, i.e.,
\begin{equation} 
 \sum_{i \neq j} SIM(\textbf{c}_i,\textbf{c}_j ).
\end{equation}
 Given the original loss $L$, e.g., least squares or cross entropy, we propose to regularize it as follows:
\begin{equation} 
L_{new}  \triangleq   L  +\beta  \sum_{i \neq j} SIM(\textbf{c}_i,\textbf{c}_j ),
\label{loss_eq}
\end{equation}
where $L_{new}$ is the augmented loss and $\beta$ is a hyper-parameter employed to control the contribution of the supplementary regularizer in the global loss of the model. The computation of the total loss is illustrated in Figure \ref{illustration_loss}.

Setting $\beta=0$ corresponds to the standard \ac{BoF} case, while a higher $\beta$ yields a loss dominated by the regularizer. In the training phase, at each step in the back-propagation, the gradient of the loss w.r.t. the parameters is computed. The additional term depends only on the elements of the dictionary, i.e., $\{\textbf{c}_1, \cdots, \textbf{c}_K\}$, of the \ac{BoF} layer. Thus, the gradient of all the model parameters, except $\{\textbf{c}_1, \cdots, \textbf{c}_K\}$, remains the same as in the standard \ac{BoF} case. For each of dictionary parameters  $\textbf{c}_i$, we have an additional feedback term equal to
\begin{equation*} 
\beta \dfrac{\partial \sum_{i \neq j} SIM(\textbf{c}_i,\textbf{c}_j ) }{\partial \textbf{c}_i},
\end{equation*}
which encourages different elements within this dictionary to be distinct. 

The proposed approach affects only the training loss and does not require any additional parameters. It can be integrated in a plug-and-play manner in any \ac{BoF}-based model to improve performance. Intuitively, the proposed schema acts as a penalty on top of the learned dictionary to provide supplementary feedback in the training phase to lessen the correlations of the codebook's items. By explicitly forcing the \ac{BoF} layer to learn a diverse and rich dictionary, we can increase the model efficiency and reach a high performance with a minimal number of parameters. 


\section{Experimental Results} \label{experimentalresults}
In this Section, we present the empirical results of the proposed approach along with the competing methods. 

\subsection{Experiment Setup}

\subsubsection{Datasets} We evaluate the performance of our approach using three different dataset:
\begin{itemize}
\item MNIST \cite{lecun1998gradient} is a dataset of $28 \times 28$ images from 10 classes.  It contains 50,000 samples for training and 10,000 for testing.
\item fashionMNIST \cite{xiao2017fashion} is a clothes dataset containing $28 \times 28$ images from 10 classes.  It has in total 50,000 samples for training and 10,000 for testing.
\item CIFAR10 \cite{krizhevsky2009learning} is an RGB image dataset containing $32\times32$ images from a total of  10 distinct classes. It has a total of 50,000 and 10,000 samples for training and testing, respectively.
\end{itemize}

\subsubsection{Training \& Testing}

In all our experiments, we hold $20\%$ of the training data for validation and hyper-parameter selection.  We also experiment with different values for the number of filters in the last convolutional layer. The full topology of the \ac{CNN} models used in MNIST/fashionMNIST and CIFAR10 experiments are reported in Table \ref{model_mnist} and Table \ref{model_cifar}, respectively. 

For MNIST and fashionMNIST experiments, all the models are trained for 50 epochs using Adam \cite{kingma2014adam} regularizer with a 0.001 learning rate and a batch-size of 128. For CIFAR10  experiments,  all the models are trained for 200 epochs with standard data augmentation \cite{he2016deep} using Adam regularizer with a 0.0001 learning rate and a batch-size of 128.

We report the competitive results of the different pooling strategies, namely global max pooling (GMP) \cite{goodfellow2016deep}, global average pooling (GAP) \cite{goodfellow2016deep}, \ac{BoF} \cite{bof,bofp}, and our approach. The size of the codebook is a hyperparameter for both \ac{BoF} and our approach. It is optimized with the validation set from $\{ 8,16,32,64,128\}$ for MNIST and fashionMNIST and from  $\{ 32,64,128,256\}$ for CIFAR10. The hyper-parameter $\beta$ in Eq. \eqref{loss_eq}, used for controlling the contribution of the proposed regularizer in the global loss, is selected from $\{ 0.1,0.01,0.001,0.0001 \}$ using the validation set in all experiments. 

\begin{table}[h]
\centering
\begin{tabular}{l}
Input layer\\
\toprule
$3\times 3 \times 32$ - Relu  \\
$3\times 3 \times 32$ - Relu  \\
$2\times 2$  max pooling layer  \\
$3\times 3 \times C$ - Relu  \\
Pooling strategy \\
Dropout (0.2) \\
512-Fully connected - Relu  \\
Dropout (0.2) \\
10-fully connected   \\
softmax layer \\
\bottomrule
\end{tabular}
\caption{Topology used for MNIST and fashionMNIST experiments. We experiment with different values of C, i.e., the number of filters in the last convolutional layer. Pooling strategy refers to the used method, e.g., global max pooling or \ac{BoF}.}
\label{model_mnist}
\end{table}

\begin{table}[h]
\centering
\begin{tabular}{l}
Input layer\\
\toprule
$3\times 3 \times 128$ - Relu  \\
$3\times 3 \times 128$ - Relu  \\
$2\times 2$  max pooling layer  \\
$3\times 3 \times 64$ - Relu  \\
$3\times 3 \times 64$ - Relu  \\
$2\times 2$  max pooling layer  \\
$3\times 3 \times C$ - Relu  \\
Pooling strategy \\
Dropout (0.2) \\
512-Fully connected - Relu  \\
Dropout (0.2) \\
10-fully connected  \\
softmax layer \\
\bottomrule
\end{tabular}
\caption{Topology used for CIFAR10 experiments. We experiment with different values of C, i.e., the number of filters in the last convolutional layer. Pooling strategy refers  to the used method, e.g., global max pooling or \ac{BoF}.}
\label{model_cifar}
\end{table}

 \begin{table*}[h]
\centering
\begin{tabular}{lcccc}
method &  16 filters &  32 filters &  64 filters  &  128 filters\\
\toprule
GMP  &3.63 $\pm$ 0.31 &  1.97 $\pm$ 0.20 & 1.39 $\pm$ 0.07 &  \textbf{1.09 $\pm$ 0.08}  \\
GAP  &4.67 $\pm$ 1.17  & 2.01 $\pm$ 0.09 & 1.31 $\pm$ 0.05 & \textbf{1.06  $\pm$ 0.02} \\
\midrule
\ac{BoF}  &  1.03 $\pm$ 0.08 &  \textbf{0.97 $\pm$ 0.11} & 1.00 $\pm$ 0.08 & 1.03 $\pm$ 0.06 \\ 
\midrule
\ac{BoF} (ours) & {1.00 $\pm$ 0.06} & 0.98 $\pm$0.06  & \underline{\textbf{0.87 $\pm$0.10}}& {0.98 $\pm$ 0.08}  \\ 
\bottomrule
\end{tabular}
\caption{Average error rates and standard deviation of different  approaches for different number of filters in the last convolutional layer on the MNIST dataset. Results are averaged over 5 random seeds. Top results for each approach are in bold and best global result is underlined. }
\label{tab:mnist_res}
\end{table*}

\subsection{ Empirical Results}

In Table \ref{tab:mnist_res} and Table \ref{tab:fmnist_res}, we report the average error rates and standard deviations for the different filter sizes, i.e., C in Table \ref{model_mnist}, on MNIST and fashionMNIST datasets, respectively. Compared to standard pooling approaches, i.e., GMP and GAP, we note that both variants of \ac{BoF} consistently yield a better performance. For the 16 filter case, for example, GMP and GAP reach $3.63\%$ and $4.67\%$ errors on MNIST, respectively, whereas standard \ac{BoF} and our variant of \ac{BoF} reach  $1.03$ and $1.00\%$ for the same case, respectively. 

\begin{table*}[h]
\centering
\begin{tabular}{lcccc}
method &  16 filters &  32 filters &  64 filters  &  128 filters \\
\toprule
GMP  & 14.94 $\pm$ 0.70 & 12.13 $\pm$ 0.30  &10.46 $\pm$ 0.18  & \textbf{9.48  $\pm$ 0.12}  \\
GAP & 15.09  $\pm$ 0.19  & 12.30 $\pm$ 0.20 & 10.91 $\pm$ 0.21 &   \textbf{9.97 $\pm$ 0.06}  \\      
\midrule
\ac{BoF}  & 9.55 $\pm$ 0.29 &  9.44 $\pm$ 0.25 & 9.04 $\pm$  0.22 &   \textbf{9.02 $\pm$ 0.15}  \\ 
\midrule
\ac{BoF} (ours) &{ 9.52 $\pm$ 0.29} & {9.14 $\pm$ 0.12} & {8.98 $\pm$ 0.18} &  \underline{\textbf{8.77 $\pm$ 0.22}} \\
\bottomrule
\end{tabular}
\caption{Average error rates and standard deviation of different  approaches for different number of filters in the last convolutional layer on the fashionMNIST dataset. Results are averaged over 5 random seeds. Top results for each approach are in bold and best global result is underlined. }
\label{tab:fmnist_res}
\end{table*}

\begin{table}[h]
\centering
\begin{tabular}{lccc}
method &   32 filters &  64 filters  &  128 filters \\
\toprule
GMP  &   22.31 $\pm$ 0.48 & 20.33 $\pm$ 0.67 & \textbf{18.80 $\pm$ 1.03}  \\
GAP   &  20.94 $\pm$ 0.53 & 20.08 $\pm$ 0.99 & \textbf{17.61 $\pm$ 0.33}  \\
\midrule
\ac{BoF}  &  {17.15 $\pm$ 0.58} & 17.05 $\pm$ 0.11 & \textbf{16.10 $\pm$ 0.20} \\
\midrule
\ac{BoF} (ours)   &   17.21 $\pm$ 0.71  & {16.57 $\pm$ 0.25}  &  \underline{\textbf{15.93 $\pm$ 0.11}}  \\
\bottomrule
\end{tabular}
\caption{Average error rates and standard deviation of different  approaches for different number of filters in the last convolutional layer on the 
CIFAR10 dataset. Results are averaged over three random seeds.}
\label{tab:cifar_res}
\end{table}

For the fashionMNIST dataset with a 16-filter model, using \ac{BoF} reduces the error rates by more than $5\%$. Compared to the standard \ac{BoF}, we note that penalizing correlations between the items of the dictionary yields in a consistently better performance in most cases. For example, for fashionMNIST using a 128-filters model, our approach reaches only $8.77\%$ error rate compared to $9.02\%$ reached by \ac{BoF}. 

For MNIST dataset, as shown in Table \ref{tab:mnist_res}, the best performances reached by GMP and GAP pooling are $1.09\%$ and $1.06\%$, respectively. In both cases, it is achieved by the 128-filters model. We note that our approach requires only 16 filters to achieve a better performance ($1.00\%$). This finding is also in agreement with the results on fashionMNIST dataset in Table \ref{tab:fmnist_res}. In this case, our approach with only 16 filters achieves better results compared to the best GAP case and only 32 filters to achieve better results compared to the best GMP case. The best performance both on MNIST and fashionMNIST is  achieved by our approach with the 64-filters and 128-filters model, respectively.



In Table \ref{tab:cifar_res}, we report the results of the different approaches on CIFAR10 dataset with three different filter sizes in the final convolutional layer, namely 32, 64 and 128 filters. As can been seen, the best result is $15.93\%$ error rate which is achieved by our approach using 128 filters. This constitutes an improvement by  $2.87$, $1.68\%$, and $0.17\%$ compared to the best results achieved by GMP, GAP, and the standard \ac{BoF}, respectively. 

\section{Conclusion and Future Work} \label{concfut}
In this paper, we proposed a scheme that builds on top of \ac{BoF} pooling to improve its performance. We proposed a regularizer, based on the pair-wise correlation of the items of the dictionary, which ensures the diversity and the richness of the learned dictionary within the \ac{BoF} layer. It led to an efficient variant of \ac{BoF} and further improved its capability, without any additional parameters. The proposed approach can be incorporated in any \ac{BoF}-based model in a plug-and-play manner. Empirical results over three different dataset showed that the proposed regularizer boosts the performance of the model and led to lower error rates. 

Future directions include more extensive experimental evaluation of the proposed approach over larger datasets and proposing more advanced techniques for quantifying the similarities between the codebook elements.


\begin{thebibliography}{10}

\bibitem{lecun2015deep}
Yann LeCun, Yoshua Bengio, and Geoffrey Hinton,
\newblock ``Deep learning,''
\newblock {\em nature}, vol. 521, no. 7553, pp. 436--444, 2015.

\bibitem{girshick2015fast}
Ross Girshick,
\newblock ``Fast r-cnn,''
\newblock in {\em Proceedings of the IEEE international conference on computer
  vision}, 2015, pp. 1440--1448.

\bibitem{krizhevsky2012imagenet}
Alex Krizhevsky, Ilya Sutskever, and Geoffrey~E Hinton,
\newblock ``Imagenet classification with deep convolutional neural networks,''
\newblock {\em Advances in neural information processing systems}, vol. 25,
  2012.

\bibitem{he2016deep}
Kaiming He, Xiangyu Zhang, Shaoqing Ren, and Jian Sun,
\newblock ``Deep residual learning for image recognition,''
\newblock in {\em Proceedings of the IEEE conference on computer vision and
  pattern recognition}, 2016, pp. 770--778.

\bibitem{simonyan2014very}
Karen Simonyan and Andrew Zisserman,
\newblock ``Very deep convolutional networks for large-scale image
  recognition,''
\newblock {\em arXiv preprint arXiv:1409.1556}, 2014.

\bibitem{ren2015faster}
Shaoqing Ren, Kaiming He, Ross Girshick, and Jian Sun,
\newblock ``Faster r-cnn: Towards real-time object detection with region
  proposal networks,''
\newblock {\em Advances in neural information processing systems}, vol. 28,
  2015.

\bibitem{zhao2019object}
Zhong-Qiu Zhao, Peng Zheng, Shou-tao Xu, and Xindong Wu,
\newblock ``Object detection with deep learning: A review,''
\newblock {\em IEEE transactions on neural networks and learning systems}, vol.
  30, no. 11, pp. 3212--3232, 2019.

\bibitem{tan2020efficientdet}
Mingxing Tan, Ruoming Pang, and Quoc~V Le,
\newblock ``Efficientdet: Scalable and efficient object detection,''
\newblock in {\em Proceedings of the IEEE/CVF conference on computer vision and
  pattern recognition}, 2020, pp. 10781--10790.

\bibitem{pang2021deep}
Guansong Pang, Longbing Cao, and Charu Aggarwal,
\newblock ``Deep learning for anomaly detection: Challenges, methods, and
  opportunities,''
\newblock in {\em Proceedings of the 14th ACM International Conference on Web
  Search and Data Mining}, 2021, pp. 1127--1130.

\bibitem{ruff2018deep}
Lukas Ruff, Robert Vandermeulen, Nico Goernitz, Lucas Deecke, Shoaib~Ahmed
  Siddiqui, Alexander Binder, Emmanuel M{\"u}ller, and Marius Kloft,
\newblock ``Deep one-class classification,''
\newblock in {\em International conference on machine learning}. PMLR, 2018,
  pp. 4393--4402.

\bibitem{park2020learning}
Hyunjong Park, Jongyoun Noh, and Bumsub Ham,
\newblock ``Learning memory-guided normality for anomaly detection,''
\newblock in {\em Proceedings of the IEEE/CVF Conference on Computer Vision and
  Pattern Recognition}, 2020, pp. 14372--14381.

\bibitem{kim2018encoding}
Taejoon Kim, Sang~C Suh, Hyunjoo Kim, Jonghyun Kim, and Jinoh Kim,
\newblock ``An encoding technique for cnn-based network anomaly detection,''
\newblock in {\em 2018 IEEE International Conference on Big Data (Big Data)}.
  IEEE, 2018, pp. 2960--2965.

\bibitem{8763885}
Jiasi Chen and Xukan Ran,
\newblock ``Deep learning with edge computing: A review,''
\newblock {\em Proceedings of the IEEE}, vol. 107, no. 8, pp. 1655--1674, 2019.

\bibitem{laakom2019color}
Firas Laakom, Jenni Raitoharju, Alexandros Iosifidis, Jarno Nikkanen, and
  Moncef Gabbouj,
\newblock ``Color constancy convolutional autoencoder,''
\newblock in {\em 2019 IEEE Symposium Series on Computational Intelligence
  (SSCI)}. IEEE, 2019, pp. 1085--1090.

\bibitem{goodfellow2016deep}
Ian Goodfellow, Yoshua Bengio, and Aaron Courville,
\newblock {\em Deep learning},
\newblock MIT press, 2016.

\bibitem{Li_2019_CVPR}
Yuchao Li, Shaohui Lin, Baochang Zhang, Jianzhuang Liu, David Doermann,
  Yongjian Wu, Feiyue Huang, and Rongrong Ji,
\newblock ``Exploiting kernel sparsity and entropy for interpretable cnn
  compression,''
\newblock in {\em Proceedings of the IEEE/CVF Conference on Computer Vision and
  Pattern Recognition (CVPR)}, June 2019.

\bibitem{lin2016towards}
Shaohui Lin, Rongrong Ji, Xiaowei Guo, Xuelong Li, et~al.,
\newblock ``Towards convolutional neural networks compression via global error
  reconstruction.,''
\newblock in {\em IJCAI}, 2016, pp. 1753--1759.

\bibitem{bof}
Nikolaos Passalis and Anastasios Tefas,
\newblock ``Neural bag-of-features learning,''
\newblock {\em Pattern Recognition}, pp. 277--294, 2017.

\bibitem{bofp}
Nikolaos Passalis and Anastasios Tefas,
\newblock ``Learning bag-of-features pooling for deep convolutional neural
  networks,''
\newblock in {\em IEEE International Conference on Computer Vision}, 2017.

\bibitem{menghani2021efficient}
Gaurav Menghani,
\newblock ``Efficient deep learning: A survey on making deep learning models
  smaller, faster, and better,''
\newblock {\em arXiv preprint arXiv:2106.08962}, 2021.

\bibitem{boforigi}
Fei-Fei Li and Pietro Perona,
\newblock ``A bayesian hierarchical model for learning natural scene
  categories,''
\newblock in {\em IEEE Computer Society Conference on Computer Vision and
  Pattern Recognition}, 2005, pp. 524--531.

\bibitem{boforigi2}
G.~Qiu,
\newblock ``Indexing chromatic and achromatic patterns for content-based colour
  image retrieval,''
\newblock {\em Pattern Recognition}, pp. 1675 -- 1686, 2002.

\bibitem{yang2007evaluating}
Jun Yang, Yu-Gang Jiang, Alexander~G Hauptmann, and Chong-Wah Ngo,
\newblock ``Evaluating bag-of-visual-words representations in scene
  classification,''
\newblock in {\em Proceedings of the international workshop on Workshop on
  multimedia information retrieval}, 2007, pp. 197--206.

\bibitem{csurka2010fisher}
Gabriela Csurka and Florent Perronnin,
\newblock ``Fisher vectors: Beyond bag-of-visual-words image representations,''
\newblock in {\em International Conference on Computer Vision, Imaging and
  Computer Graphics}. Springer, 2010, pp. 28--42.

\bibitem{shukla2022bag}
Jyoti~S Shukla, Kriti Rastogi, Hetal Patel, Gaurav Jain, and Shashikant Sharma,
\newblock ``Bag of visual words methodology in remote sensing—a review,''
\newblock in {\em Proceedings of the International e-Conference on Intelligent
  Systems and Signal processing}. Springer, 2022, pp. 475--486.

\bibitem{zhang2010understanding}
Yin Zhang, Rong Jin, and Zhi-Hua Zhou,
\newblock ``Understanding bag-of-words model: a statistical framework,''
\newblock {\em International journal of machine learning and cybernetics}, vol.
  1, no. 1, pp. 43--52, 2010.

\bibitem{passalis2018training}
Nikolaos Passalis and Anastasios Tefas,
\newblock ``Training lightweight deep convolutional neural networks using
  bag-of-features pooling,''
\newblock {\em IEEE transactions on neural networks and learning systems}, vol.
  30, no. 6, pp. 1705--1715, 2018.

\bibitem{chumachenko2022self}
Kateryna Chumachenko, Alexandros Iosifidis, and Moncef Gabbouj,
\newblock ``Self-attention neural bag-of-features,''
\newblock {\em arXiv preprint arXiv:2201.11092}, 2022.

\bibitem{8487014}
Nikolaos Passalis, Anastasios Tefas, Juho Kanniainen, Moncef Gabbouj, and
  Alexandros Iosifidis,
\newblock ``Temporal bag-of-features learning for predicting mid price
  movements using high frequency limit order book data,''
\newblock {\em IEEE Transactions on Emerging Topics in Computational
  Intelligence}, vol. 4, no. 6, pp. 774--785, 2020.

\bibitem{bof3}
Alexandros Iosifidis, Anastastios Tefas, and Ioannis Pitas,
\newblock ``Discriminant bag of words based representation for human action
  recognition,''
\newblock {\em Pattern Recognition Letters}, pp. 185 -- 192, 2014.

\bibitem{passalis2020efficient}
Nikolaos Passalis, Jenni Raitoharju, Anastasios Tefas, and Moncef Gabbouj,
\newblock ``Efficient adaptive inference for deep convolutional neural networks
  using hierarchical early exits,''
\newblock {\em Pattern Recognition}, vol. 105, pp. 107346, 2020.

\bibitem{tran2020attention}
Dat~Thanh Tran, Nikolaos Passalis, Anastasios Tefas, Moncef Gabbouj, and
  Alexandros Iosifidis,
\newblock ``Attention-based neural bag-of-features learning for sequence
  data,''
\newblock {\em arXiv preprint arXiv:2005.12250}, 2020.

\bibitem{bof2}
Yu-Gang Jiang, Chong-Wah Ngo, and Jun Yang,
\newblock ``Towards optimal bag-of-features for object categorization and
  semantic video retrieval,''
\newblock in {\em ACM International Conference on Image and Video Retrieval},
  2007, pp. 494--501.

\bibitem{laakom2020bag}
Firas Laakom, Nikolaos Passalis, Jenni Raitoharju, Jarno Nikkanen, Anastasios
  Tefas, Alexandros Iosifidis, and Moncef Gabbouj,
\newblock ``Bag of color features for color constancy,''
\newblock {\em IEEE Transactions on Image Processing}, vol. 29, pp. 7722--7734,
  2020.


\bibitem{chen2008diversity}
Huanhuan Chen,
\newblock {\em Diversity and regularization in neural network ensembles},
\newblock Ph.D. thesis, University of Birmingham, 2008.

\bibitem{gangeh2015supervised}
Mehrdad~J Gangeh, Ahmed~K Farahat, Ali Ghodsi, and Mohamed~S Kamel,
\newblock ``Supervised dictionary learning and sparse representation-a
  review,''
\newblock {\em arXiv preprint arXiv:1502.05928}, 2015.

\bibitem{rajesh2018heuristic}
K~Rajesh and Atul Negi,
\newblock ``Heuristic based learning of parameters for dictionaries in sparse
  representations,''
\newblock in {\em 2018 IEEE Symposium Series on Computational Intelligence
  (SSCI)}. IEEE, 2018, pp. 1013--1019.

\bibitem{ayinde2019regularizing}
Babajide~O Ayinde, Tamer Inanc, and Jacek~M Zurada,
\newblock ``Regularizing deep neural networks by enhancing diversity in feature
  extraction,''
\newblock {\em IEEE transactions on neural networks and learning systems}, vol.
  30, no. 9, pp. 2650--2661, 2019.

\bibitem{naderahmadian2015correlation}
Yashar Naderahmadian, Soosan Beheshti, and Mohammad~Ali Tinati,
\newblock ``Correlation based online dictionary learning algorithm,''
\newblock {\em IEEE Transactions on signal processing}, vol. 64, no. 3, pp.
  592--602, 2015.

\bibitem{van2012kernel}
Hien Van~Nguyen, Vishal~M Patel, Nasser~M Nasrabadi, and Rama Chellappa,
\newblock ``Kernel dictionary learning,''
\newblock in {\em 2012 IEEE International Conference on Acoustics, Speech and
  Signal Processing (ICASSP)}. IEEE, 2012, pp. 2021--2024.

\bibitem{jing2014uncorrelated}
Xiao-Yuan Jing, Rui-Min Hu, Fei Wu, Xi-Lin Chen, Qian Liu, and Yong-Fang Yao,
\newblock ``Uncorrelated multi-view discrimination dictionary learning for
  recognition,''
\newblock in {\em Proceedings of the AAAI Conference on Artificial
  Intelligence}, 2014, vol.~28.

\bibitem{wang2020orthogonal}
Jiayun Wang, Yubei Chen, Rudrasis Chakraborty, and Stella~X Yu,
\newblock ``Orthogonal convolutional neural networks,''
\newblock in {\em Proceedings of the IEEE/CVF conference on computer vision and
  pattern recognition}, 2020, pp. 11505--11515.

\bibitem{zbontar2021barlow}
Jure Zbontar, Li~Jing, Ishan Misra, Yann LeCun, and St{\'e}phane Deny,
\newblock ``Barlow twins: Self-supervised learning via redundancy reduction,''
\newblock in {\em International Conference on Machine Learning}. PMLR, 2021,
  pp. 12310--12320.


\bibitem{pmlrxie17b}
Pengtao Xie, Aarti Singh, and Eric~P. Xing,
\newblock ``Uncorrelation and evenness: a new diversity-promoting
  regularizer,''
\newblock in {\em Proceedings of the 34th International Conference on Machine
  Learning}, 2017, pp. 3811--3820.

\bibitem{cogswell2015reducing}
Michael Cogswell, Faruk Ahmed, Ross Girshick, Larry Zitnick, and Dhruv Batra,
\newblock ``Reducing overfitting in deep networks by decorrelating
  representations,''
\newblock {\em arXiv preprint arXiv:1511.06068}, 2015.

\bibitem{laakom2021feature}
Firas Laakom, Jenni Raitoharju, Alexandros Iosifidis, and Moncef Gabbouj,
\newblock ``On feature diversity in energy-based models,''
\newblock in {\em Energy Based Models Workshop-ICLR 2021}, 2021.

\bibitem{laakom2022reducing}
Firas Laakom, Jenni Raitoharju, Alexandros Iosifidis, and Moncef Gabbouj,
\newblock ``Reducing redundancy in the bottleneck representation of the
  autoencoders,''
\newblock {\em arXiv preprint arXiv:2202.04629}, 2022.

\bibitem{laakom2021within}
Firas Laakom, Jenni Raitoharju, Alexandros Iosifidis, and Moncef Gabbouj,
\newblock ``Within-layer diversity reduces generalization gap,''
\newblock {\em arXiv preprint arXiv:2106.06012}, 2021.

\bibitem{lecun1998gradient}
Yann LeCun, L{\'e}on Bottou, Yoshua Bengio, and Patrick Haffner,
\newblock ``Gradient-based learning applied to document recognition,''
\newblock {\em Proceedings of the IEEE}, vol. 86, no. 11, pp. 2278--2324, 1998.

\bibitem{xiao2017fashion}
Han Xiao, Kashif Rasul, and Roland Vollgraf,
\newblock ``Fashion-mnist: a novel image dataset for benchmarking machine
  learning algorithms,''
\newblock {\em arXiv preprint arXiv:1708.07747}, 2017.

\bibitem{krizhevsky2009learning}
Alex Krizhevsky, Geoffrey Hinton, et~al.,
\newblock ``Learning multiple layers of features from tiny images,''
\newblock 2009.

\bibitem{kingma2014adam}
Diederik~P Kingma and Jimmy Ba,
\newblock ``Adam: A method for stochastic optimization,''
\newblock {\em arXiv preprint arXiv:1412.6980}, 2014.

\end{thebibliography}
\end{document}